\documentclass[conference]{IEEEtran}
\IEEEoverridecommandlockouts

\pdfoutput=1

\usepackage{cite}
\usepackage{amsmath}
\usepackage{amssymb}
\usepackage{booktabs}
\usepackage{graphicx}
\usepackage{multirow}
\usepackage{url}
\usepackage[hidelinks]{hyperref}
\usepackage{placeins}
\usepackage[T1]{fontenc}
\usepackage[utf8]{inputenc}
\usepackage{tikz}
\usetikzlibrary{arrows.meta,calc,fit,positioning,shapes.geometric}
\usepackage{tabulary}
\usepackage{array}
\usepackage{makecell}
\usepackage{ragged2e}
\usepackage{multicol}

\AtBeginDocument{%
  \setlength{\abovedisplayskip}{2pt plus 1pt minus 1pt}%
  \setlength{\belowdisplayskip}{2pt plus 1pt minus 1pt}%
  \setlength{\abovedisplayshortskip}{1pt plus 1pt}%
  \setlength{\belowdisplayshortskip}{2pt plus 1pt minus 1pt}%
}
\newcolumntype{P}[1]{>{\Centering\arraybackslash}p{#1}}

\begin{document}

\title{FinsSim: A Reality-Aligned Integrated Simulation Platform for Underwater Robot Learning}

\author{Yu Zhang$^{1}$, Yuanmingqing Song$^{2}$, Xiangyun Rao$^{1}$,\\ 
Pangkit Fong$^{1}$, Kunhao Zhang$^{3}$, Chongrong Fang$^{1}$ and Jianping He$^{1*}$%
\thanks{$^{1}$Yu Zhang, Xiangyun Rao, Pangkit Fong, Chongrong Fang and Jianping He are with the Department of Automation, Shanghai Jiao Tong University, Shanghai, China
{\tt\small \{sherlock\_nolan, Raoxiangyun, fpjgaoge, crfang, jphe\}@sjtu.edu.cn}}%
\thanks{$^{2}$Yuanmingqing Song is with the School of Ocean and Civil Engineering, Shanghai Jiao Tong University, Shanghai, China
{\tt\small symq20060606@sjtu.edu.cn}}%
\thanks{$^{3}$Kunhao Zhang is with the Zhejiang University, Hangzhou, China
{\tt\small 3220105256@zju.edu.cn}}%
}

\maketitle
\thispagestyle{empty}
\pagestyle{empty}

\begin{abstract}
Underwater robot learning relies on simulators that integrate high-fidelity hydrodynamics, convenient learning interfaces, and a credible transition to real scenarios.
In this work, we present FinsSim, a reality-aligned integrated simulation platform for Sim‑to‑Real underwater robot learning.
FinsSim first constructs high-fidelity simulation with selectable backends to adapt to diverse requirements.
To facilitate underwater robot research, it further offers standard control baselines, alongside with unified robot learning workflows.
For reliable Sim-to-Real transfer, FinsSim adopts a multi-sensor fusion scheme to provide low-cost yet precise localization. Moreover, it implements calibrated thruster-hydrodynamics models and a constrained wrench allocation algorithm.
Bridging these modules by ROS~2, FinsSim establishes a complete Sim-to-Real transfer pipeline.
Through matched simulations and experiments, it is demonstrated that reliable Sim-to-Real transfer of underwater robot control policies can be achieved with the FinsSim framework. Separate ablation studies also validate that the modules of FinsSim can address the pivotal issues of underwater Sim-to-Real from different aspects.
Overall, this work aims to bridge the gap between theoretical research and practical applications, ultimately driving advancements in the field of underwater robotics.
\end{abstract}

\begin{IEEEkeywords}
Underwater Robotics, Simulation Platform, Sim-to-Real, Robot Learning
\end{IEEEkeywords}

\section{Introduction}

The dynamics of underwater robots are affected by various factors, such as uncertain hydrodynamic coefficients,
nonlinear damping, and environmental disturbances\cite{aldhaheri2025underwater}.
Underwater robot learning has emerged as an appealing approach for control of underwater robot for its convenience \cite{Hong2026DataDrivenMarineControl}.
However, collecting sufficient interaction data on real scenarios is costly, time-consuming, and even potentially unsafe. Simulators therefore play a central role in underwater robot learning. Providing a high-fidelity environment, simulators empower policy training for underwater robot and large-scale evaluation before deployment to real scenarios
\cite{Cai2025LearningToSwim,Chu2025MarineGym}.

\begin{table*}[t]
\caption{Platform scope reported by the cited primary publications.}
\label{tab:platform_scope}
\centering
\footnotesize
\setlength{\tabcolsep}{3pt}
\renewcommand{\arraystretch}{1.15}

\begin{tabulary}{\textwidth}{@{}
  P{0.105\textwidth}
  P{0.035\textwidth}
  C
  C
  P{0.075\textwidth}
  C
  P{0.07\textwidth}
  P{0.09\textwidth}
@{}}
\toprule
\makecell{Platform} &
\makecell{MA} &
\makecell{Backend} &
\makecell{Hydrodynamics} &
\makecell{ROS} &
\makecell{Learning\\Interface} &
\makecell{Training\\Parallelism} &
\makecell{Sim-to-Real\\evidence} \\
\midrule
\makecell[c]{UUV \\Simulator~\cite{Manhaes2016UUVSimulator}}
& $\times$
& Gazebo Classic
& Fossen-form dynamics
& ROS 1
& $\times$
& $\times$
& -- \\

DAVE~\cite{Zhang2022DAVE}
& $\times$
& Gazebo Classic
& Fossen-form dynamics
& ROS 1
& $\times$
& $\times$
& -- \\

HoloOcean~\cite{Romrell2025HoloOcean2}
& $\checkmark$
& Unreal Engine
& Fossen-form dynamics
& ROS 2
& Python
& $\times$
& -- \\

Stonefish~\cite{Grimaldi2025Stonefish}
& $\times$
& Bullet/OpenGL
& Mesh-based hydrodynamics
& ROS~1/2
& $\times$
& $\times$
& -- \\

MARUS~\cite{Loncar2022MARUS}
& $\checkmark$
& Unity
& Mesh-based hydrodynamics (Simplified)
& ROS~1/2
& $\times$
& $\times$
& -- \\

MarineGym~\cite{Chu2025MarineGym}
& $\checkmark$
& Isaac Lab
& Fossen-form dynamics (Simplified)
& $\times$
& TorchRL/TensorDict
& $\checkmark$
& -- \\

UNav-Sim~\cite{Amer2023UNavSim}
& $\times$
& UE5/AirSim
& Fossen-form dynamics
& ROS~1/2
& AirSim Gym / SB3
& $\times$
& -- \\

Orca~\cite{McQueen2022Orca4,McQueen2026Orca5}
& $\times$
& \makecell[c]{Gazebo Harmonic/\\ArduSub SITL}
& Fossen-form dynamics (Simplified)
& ROS~2
& $\times$
& $\times$
& -- \\

FinsSim (ours)
& $\checkmark$
& Unity/Isaac Lab
& \makecell[c]{Fossen-form dynamics (Simplified)/\\Fossen-form dynamics (Calibrated)/\\ Mesh-based hydrodynamics}
& ROS~2
& \makecell[c]{Python/Gym-like;\\ TorchRL/TensorDict;\\
SB3/Imitation/ MARL}
& $\checkmark$
& $\checkmark$ \\

\bottomrule
\end{tabulary}

\vspace{3pt}
\begin{minipage}{0.98\textwidth}
\footnotesize
MA denotes multi-agent support; $\checkmark$ denotes native platform support, whereas $\times$ denotes that such support is not reported in the cited publication.
External integrations may provide capabilities not listed in the original source.
\end{minipage}
\end{table*}

Existing underwater simulators provide complementary capabilities as summarized in Table~\ref{tab:platform_scope}.
Classical marine simulators provide mature vehicle and sensor models but are not
designed towards learning-friendliness~\cite{Manhaes2016UUVSimulator,Zhang2022DAVE,Loncar2022MARUS,McQueen2022Orca4,McQueen2026Orca5,Grimaldi2025Stonefish}.
Recent learning-oriented platforms emphasize GPU-native parallel simulation,
but they do not expose a complete, calibration-traceable path from simulated actions to physical sensing and actuation \cite{Potokar2022HoloOcean,Chu2025MarineGym,Amer2023UNavSim}. Besides, to the best of our knowledge, no existing simulator provides a reusable pipeline for deploying policies learned in the simulation across vehicles, tasks and environments.
Overall, all these works have promoted the development of underwater robot learning.

However, these capabilities remain scattered across separate platforms.
Meanwhile, reproducing an underwater Sim-to-Real experiment typically requires researchers to assemble these capabilities.
As a result, most evaluations of underwater Reinforcement Learning (RL) are confined to simulation environments, while hardware demonstrations typically adopt task‑specific Sim-to-Real pipelines which are hard to reuse across different environments
\cite{Cai2025LearningToSwim,Sufan2025Swim4Real,Fosso2025Sim2Swim,Tuncay2025FastPolicyLearning,Morgan2026AdaptiveDynamics}.
For researchers, the practical barrier to underwater RL is therefore not only the absence of a simulator, but the absence of a
\emph{reality-aligned integrated simulation platform}, in which the necessary components are systematically integrated and consistently aligned with real‑world environments.

To address this gap, we present \textbf{FinsSim}, an open-source
simulation platform for underwater robot learning and Sim-to-Real transfer. 
FinsSim first constructs high-fidelity simulation with selectable backends to adapt to diverse requirements, alongside unified single- and multi-agent learning workflows.
For reliable Sim-to-Real transfer, FinsSim closes this gap by coupling 
reality-aligned model, reliable localization, and safety-constrained controllers within a unified simulation-to-hardware workflow. Finally, ROS~2 connects all the components, making it a reusable full-stack framework.
The main contributions of this work are summarized as follows:

\begin{itemize}
    \item \textbf{Fidelity-Scalable Hydrodynamics:}
    To accommodate the diverse configurations,
    FinsSim exposes parameterized and geometry-based hydrodynamic models as interchangeable simulation backends. Its hydrodynamics of simulation is suitable for different model construction methods. Meanwhile, FinsSim implement a dedicated system identification procedure for the calibration of hydrodynamic coefficients, which improves the fidelity and transferability across different scenarios.

    \item \textbf{Unified Learning Workflows:}
    To empower the development of multimodal underwater control policy,
    FinsSim supports various sensor simulations which provide multi-source information. It also provides conventional control baselines, together with single- and multi-agent training interfaces. Moreover, FinsSim integrates parallel training schemes to support GPU-acceleration.
    With FinsSim, users can conveniently conduct simulation control, data collection, and policy training.

    \item \textbf{Reliable Sim-to-Real Transfer:}
    To bridge the persistent reality gap caused by uncalibrated dynamics and unreliable underwater state estimation, FinsSim provides a tightly coupled calibration and localization method that aligns every simulated component with the physical system. Specifically,
    FinsSim first constructs a pool-scale global localization system. Despite its low cost, FinsSim still achieves fairly precise localization. FinsSim also provides the unified calibration procedures of thruster and actuator-dynamics, providing safety-constrained controllers. With above all, FinsSim ultimately aligns the simulated models with the real vehicles, realizing reliable Sim-to-Real transfer.

    \item \textbf{Complete Experimental Verification:}
    To validate the efficacy of FinsSim, we first conduct experiments on the reliability of localization and simulation dynamics.
    For Sim-to-Real transfer reliability, we evaluate station keeping and long-horizon
    trajectory tracking in matched simulation and pool experiments.
    Beyond reporting policy performance, we also study how each module of FinsSim addresses the Sim-to-Real gap. Experiments demonstrate that reliable Sim-to-Real transfer of underwater robot control policies can be achieved with FinsSim, and almost all modules of FinsSim benefit the Sim‑to‑Real transfer performance from different aspects.
\end{itemize}

\section{Related Work}
\label{sec:related}
\subsection{Underwater Simulation Platforms}

Underwater simulators primarily differ in their hydrodynamic abstraction. UUV
Simulator and DAVE use parameterized Fossen-style 6-DOF dynamics with added
mass, damping, restoring forces, and thruster models
\cite{Fossen2011Handbook,Manhaes2016UUVSimulator,Zhang2022DAVE}. HoloOcean and MarineGym expose related configurable and learning-oriented model families
\cite{Potokar2022HoloOcean,Chu2025MarineGym}. In contrast, Stonefish computes loads from dedicated physical geometry, while MARUS and UNav-Sim emphasize Unity/Unreal scenes and sensing~\cite{Grimaldi2025Stonefish,Loncar2022MARUS,Amer2023UNavSim}.
Mesh/surface models, CFD or potential-flow estimations, and empirical fitting are thus complementary tools to constructing vehicle models.
FinsSim adopts both Unity and Isaac Lab as backends, and exposes Fossen-style, surface/mesh, and simplified Fossen models. In this way, FinsSim constructs simulations that adapt to diverse requirements.

\begin{figure*}[t]
  \centering
  \includegraphics[width=1.0\textwidth]{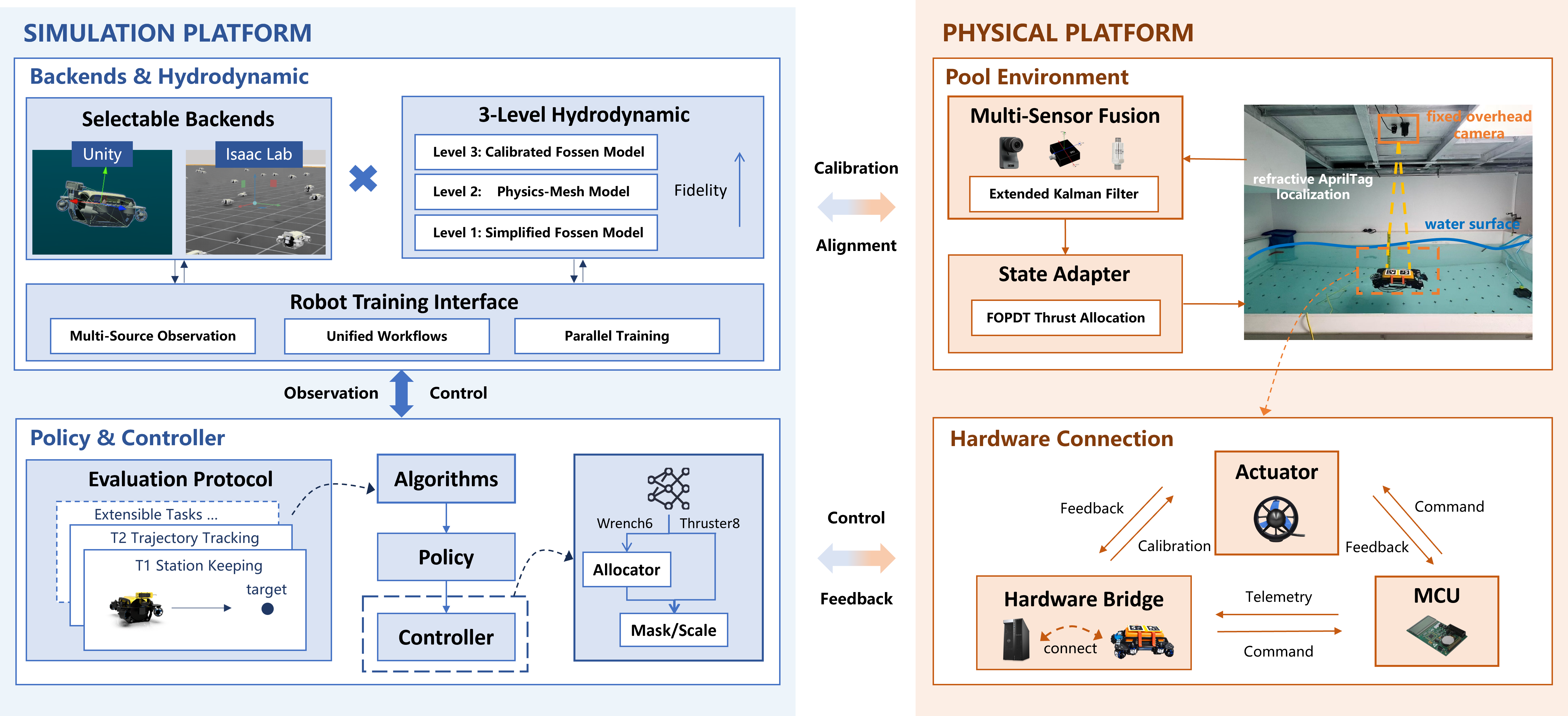}
  \caption{Overview of FinsSim.
    The platform connects Unity and Isaac Lab
    simulation with convenient learning and control interfaces and a
    hardware-connected runtime (using FinsROV~\cite{FPJGAOGE2025FinsROV} as an example).
    The physical deployment combines refractive AprilTag localization,
    state estimation, bounded actuation, ROS~2 communication, and
    calibrated eight-thruster execution.
    T1 and T2 denote the evaluated station-keeping and trajectory-tracking
    tasks, respectively.}
  \label{fig:system_architecture}
\end{figure*}

\subsection{Sim-to-Real Underwater Reinforcement Learning}

Underwater RL has been applied to 6-DOF pose with diverse tasks. Recent studies report both thruster-level policies and direct Sim-to-Real control on underwater vehicles
\cite{Cai2025LearningToSwim,Sufan2025Swim4Real,Tuncay2025FastPolicyLearning,Fosso2025Sim2Swim}.
Thruster-level actions jointly learn allocation and control but depend on a
specific layout and motor model. Instead, wrench-level actions rely on a
vehicle-specific allocator. 
During Sim-to-Real transfer,
domain randomization and adaptive dynamics are also introduced to further
address model uncertainty~\cite{Lu2023DataInformedDR,Morgan2026AdaptiveDynamics}.
Meanwhile, the control task determines what a transfer evaluation reveals. Station keeping tests steady-state regulation and disturbance rejection \cite{Chang2025LearningToDock}. Trajectory tracking
tests sustained motion, turns, reversals, and accumulated model error \cite{Chang2025LearningToDock}. Docking adds terminal precision and perception~\cite{Patil2021DockingBenchmark}.
Accordingly, we evaluate FinsSim on matched simulation and pool trials for 4-DOF station keeping and 3-dimensional trajectory tracking.
Results demonstrate the effectiveness of FinsSim on Sim-to-Real transfer.

\subsection{Underwater Localization, Calibration and Deployment}

Underwater localization usually fuses IMU and pressure depth with DVL, acoustic,
or external measurements~\cite{Kinsey2006UnderwaterNavigation}, generally with a high cost. For example, the Water Linked
Underwater GPS G2 BlueROV2 Kit~\cite{Watson2020UUVLocalization} is listed at USD~8,990, while a commercially deployed underwater
optical motion-capture system~\cite{Jia2026UWMBSM,Khanmeh2026ROVMocap} is  approximately
USD~118--266\,k. Meanwhile, camera-based
infrastructure offers a lower-cost pool-scale alternative, which must account for air-water refraction. This motivates many localization methods
\cite{Tian2009SeeingThroughWater,Suresh2019ThroughWaterSLAM,Carver2022Sunflower}.
For better deployment, it also requires calibration of buoyancy, Fossen parameters, thrust curves, and related quantities. CFD or potential-flow analysis can initialize the model to be identified, while calibration connects it to the real vehicle. For example, BlueROV2 combines a Fossen model with estimated added mass, tuned damping, and a test-facility-validated thruster model~\cite{vonBenzon2022BlueROV2}.

Inspired by above works, FinsSim uses a fixed calibrated camera and two body-mounted AprilTags
\cite{Olson2011AprilTag}. It also adopts a pressure sensor and an IMU. Fusing information from these sensors, FinsSim achieves fairly precise localization despite its low cost, for a documented hardware
expense of USD~100.
For deployment,
FinsROV~\cite{FPJGAOGE2025FinsROV} similarly uses a calibrated allocation matrix,
direction-dependent thrust and motor models. Connected by a ROS~2 hardware bridge, FinsSim establishes a complete Sim-to-Real deployment pipeline, which separates vehicle calibration from task specification. 
For a new task
on the same vehicle, perception, allocation, safety limits, and
dynamic profiles are reused, while only the task definition and policy should be
retrained. T1 and T2 demonstrate this reuse on the same calibrated FinsROV
stack. For a new vehicle, the ROS~2 interface and workflow remain reusable,
while vehicle dynamics, thruster calibration, allocation, and
safety limits must be re-established.

\section{Methodology}
\label{sec:methodology}

\subsection{Overview}
\label{sec:overview}

As Fig.~\ref{fig:system_architecture} has shown, FinsSim is divided into \textit{simulation platform} and \textit{physical platform}.
The simulation platform supports both Unity and Isaac Lab as simulation backends. Using
gRPC-ROS~2 bridge~\cite{Loncar2022MARUS}, it permits ROS~2 controllers to operate simulated vehicles
directly, and the learning interface is developed based on ML-Agents~\cite{Juliani2020Unity}.
Physical platform achieves accurate global localization via a multi-sensor fusion scheme. Using a hardware bridge to relay control commands to the on-board MCU, it enables calibrated thruster actuation on the real underwater vehicle.
Connected by ROS~2, FinsSim ultimately provides a complete Sim-to-Real closed-loop.

\subsection{Simulation Platform}
\label{sec:simulation platform}

\subsubsection{Hydrodynamic Modeling}
FinsSim provides multiple hydrodynamic models as interchangeable simulation backends. Rather than introducing a new hydrodynamic formulation, it allows users to select a suitable model according to their requirements. Specifically, FinsSim provides three hydrodynamic models: a calibrated Fossen model, a physics-mesh-based model, and a simplified Fossen model.

\textbf{Calibrated Fossen Model.} FinsSim provides a high-fidelity Fossen model through calibration.
The parameterized model uses the standard decomposition \cite{Fossen2011Handbook}, which is given by
\begin{equation}
\boldsymbol{\tau}_{\mathrm{h}}^{\mathrm{F}}=
\boldsymbol{\tau}_{\mathrm{H}}+\boldsymbol{\tau}_{\mathrm{D}}(\boldsymbol{\nu}_{\mathrm{r}})
-\mathbf{M}_{\mathrm{A}}\dot{\boldsymbol{\nu}}_{\mathrm{r}}
-\mathbf{C}_{\mathrm{A}}(\boldsymbol{\nu}_{\mathrm{r}})\boldsymbol{\nu}_{\mathrm{r}},
\label{eq:hydrodynamic_wrench}
\end{equation}
where \(\boldsymbol{\tau}_{\mathrm{H}}\) is the hydrostatic restoring wrench, \(\boldsymbol{\nu}_{\mathrm{r}}\) is the fluid-relative velocity, \(\mathbf{M}_{\mathrm{A}}\) is
the added-mass matrix, and \(\mathbf{C}_{\mathrm{A}}\) is its Coriolis/centripetal matrix. \(\boldsymbol{\tau}_{\mathrm{D}}\) is the damping term, which is implemented as
\begin{equation}
 \boldsymbol{\tau}_{\mathrm{D}}=-\left[\mathbf{D}_1+\mathbf{D}_2\operatorname{diag}
 \left(\lvert\boldsymbol{\nu}_{\mathrm{r}}\rvert\right)+\mathbf{D}_u\lvert u_{\mathrm{r}}\rvert\right]\boldsymbol{\nu}_{\mathrm{r}}.
 \label{eq:fossen_damping}
\end{equation}

The calibration procedure is described in Sec.~\ref{sec:fossen_calibration}.

\textbf{Physics-Mesh Model.}
FinsSim also supports modeling based on mesh files.
The geometry-based backend clips a closed physics mesh to enclose the vehicle and accumulates drag over the triangles beneath the water surface.
Specifically, given triangle \(i\), let \(A_i\),
\(\mathbf c_i\), and \(\mathbf n_i\) denote its area, centroid, and outward
normal. Its local relative flow and normal/tangential components are
\begin{equation}
 \mathbf v_i=\mathbf v_{\mathrm{w}}(\mathbf c_i)-\mathbf v_{\mathrm{body}}(\mathbf c_i),
 ~ \mathbf v_{t,i}=(\mathbf I-\mathbf n_i\mathbf n_i^{\mathsf T})\mathbf v_i.
 \label{eq:mesh_velocity}
\end{equation}

Then the implemented form and skin-drag forces are
\begin{equation}
\begin{aligned}
\mathbf{F}^{\mathrm{form}}_i &=
\tfrac{1}{2}\rho C_{\mathrm{form}} A_i
[-\mu_i]_+\,\mathbf{v}_i\lVert\mathbf{v}_i\rVert_2,\\
\mathbf{F}^{\mathrm{skin}}_i &=
\tfrac{1}{2}\rho C_{\mathrm{skin}}A_i
\mathbf{v}_{t,i}\lVert\mathbf{v}_{t,i}\rVert_2 .
\end{aligned}
\label{eq:mesh_face_forces}
\end{equation}
where \(C_{\mathrm{form}}\) and \(C_{\mathrm{skin}}\) are tunable coefficients, \([a]_+=\max(a,0)\) and \(\mu_i=\mathbf n_i^{\mathsf T}\mathbf v_i/\lVert\mathbf v_i\rVert_2\) is the inflow cosine.
These forces and their moments about the center of mass are summed to form the mesh wrench.
The implementation follows the
geometry-aware modeling principle used by Stonefish~\cite{Grimaldi2025Stonefish}. Commercial Unity plugin DWP2 can supply an extra option~\cite{NWH2026DWP2}.

\textbf{Simplified Fossen Model.} FinsSim provides a simplified Fossen model,
which follows the same learning-oriented motivation in~\cite{Cai2025LearningToSwim}.
The simplified backend retains hydrostatics and damping, which is given by
\begin{equation}
\boldsymbol{\tau}_{\mathrm{h}}^{\mathrm{S}}
=\boldsymbol{\tau}_{\mathrm{H}}+\boldsymbol{\tau}_{\mathrm{D}}(\boldsymbol{\nu}).
\label{eq:simplified_fossen}
\end{equation}

Therefore, it avoids relative-acceleration estimation and added-mass Coriolis
evaluation at every physics step. Utilizing DR to avoid accurate modeling, the reduced model is useful during large-batch RL where parallel training is necessary.

\subsubsection{Fossen Calibration}
\label{sec:fossen_calibration}
In the controller body frame,
let \(\boldsymbol{\nu}=[u,v,w,p,q,r]^{\mathsf T}\) denote surge, sway, heave, roll, pitch,
and yaw. For the translation axes and yaw, the identification workflow fits with the following equation.
\begin{equation}
\tau_i=m_i^{\mathrm{eff}}\dot{\nu}_i+d_{1,i}\nu_i+
d_{2,i}|\nu_i|\nu_i+b_i,
\label{eq:fossen_identification}
\end{equation}
where \(\tau_i\) is the applied force or moment, and
\(m_i^{\mathrm{eff}}\), \(d_{1,i}\), \(d_{2,i}\), and \(b_i\) denote the fitted
effective mass or inertia, linear damping, quadratic damping, and constant bias,
respectively.

Specifically, roll and pitch use the analogous effective-inertia model with the gravity-buoyancy restoring moment term \(k_i\sin\eta_i\), where \(\eta_i\in\{\phi,\theta\}\). Single-axis step trials excite surge, sway, and yaw. For heave, we use a dive-and-coast sequence to expose both powered descent and natural ascent.

\subsubsection{Sensing \& Learning}
Following MARUS~\cite{Loncar2022MARUS}, FinsSim supports camera, IMU, and depth-sensor models.
Visual observations are generated using Unity's High Definition Render Pipeline
(HDRP) water and rendering stack, which provides a
realistic visual observation source. These sensor models allow robot policies
to use either state-based or visual observations.

Moreover, FinsSim develops conventional PID control baselines for ROV control tasks, together with RL, IL, and MARL training interfaces. It also provides a convenient DR toolkit, including randomizing body mass and volume, hydrodynamic coefficients, thruster characteristics, etc. For parallel training, Unity utilizes ML-Agents for parallel areas while Isaac Lab provides a GPU-native batched implementation for high-throughput state-based tasks~\cite{Juliani2020Unity,Mittal2025IsaacLab}.
With FinsSim, users can conveniently conduct simulation control, data collection, and policy training.

\subsection{Physical Platform}
\label{sec:Physical}

FinsSim closes the simulation-to-real loop through calibrated perception, actuation, and ROS~2 control modules, which are specified in the following text.

\subsubsection{Localization}
Rather than requiring an acoustic infrastructure or a vehicle-borne DVL, FinsSim uses a fixed overhead camera, a pressure sensor, an IMU and AprilTags on board to provide a
low-cost and easy-to-access global localization for pool experiments.
Specifically, refractive
AprilTag geometry estimates the global horizontal position and yaw, while a pressure sensor and IMU provide depth and roll/pitch, respectively.
Throughout this subsection, \(p\), \(b\), \(c\), and \(t\) denote the pool,
vehicle body, camera, and AprilTag frames, respectively. We use
\begin{equation}
 \mathbf T_{ab}:\ \mathcal{F}_b\rightarrow\mathcal{F}_a,
 \quad \mathbf p_a=\mathbf T_{ab}\mathbf p_b,
 \label{eq:frame_transform_convention}
\end{equation}
so that the first subscript denotes the destination frame and the second the
source frame. For rays, superscripts \(a\) and \(w\) denote the air-side and
water-side segments, respectively.

\textbf{Refractive AprilTag.}
Consider one AprilTag. The detector returns four image corners
\(\mathbf u_i=[u_i,v_i]^{\mathsf T}\), \(i\in\{0,\ldots,3\}\). After undistortion with
the camera intrinsics \(\mathbf K\) and distortion parameters \(\mathbf d\), we get the normalized image coordinate \([\bar u_i,\bar v_i]^{\mathsf T}\). The calibrated camera pose is
\(\mathbf T_{pc}=[\mathbf R_{pc},\mathbf C_p]\), where \(\mathbf R_{pc}\) rotates camera-frame vectors
to the pool frame and \(\mathbf C_p\) is the camera center. For the water interface, \(\Pi:\mathbf n_{\Pi}^{\mathsf T}\mathbf X+b_{\Pi}=0\), the pixel is then back-projected as
\begin{equation}
 \mathbf r_{c,i}=
 \frac{[\bar u_i,\,\bar v_i,\,1]^{\mathsf T}}
 {\left\lVert[\bar u_i,\,\bar v_i,\,1]^{\mathsf T}\right\rVert_2},
 \quad \mathbf r_{p,i}^{\mathrm{a}}=\mathbf R_{pc}\mathbf r_{c,i}.
 \label{eq:air_ray}
\end{equation}
where \(\mathbf r_{c,i}\) and \(\mathbf r_{p,i}^{\mathrm{a}}\) are the unit ray directions
in the camera and pool frames, respectively. The air ray
\(\mathbf C_p+s\mathbf r_{p,i}^{\mathrm{a}}\) intersects the water surface at
\begin{equation}
 s_i=-\frac{\mathbf n_{\Pi}^{\mathsf T}\mathbf C_p+b_{\Pi}}
 {\mathbf n_{\Pi}^{\mathsf T}\mathbf r_{p,i}^{\mathrm{a}}},
 \quad \mathbf S_{p,i}=\mathbf C_p+s_i\mathbf r_{p,i}^{\mathrm{a}}.
 \label{eq:surface_intersection}
\end{equation}
where \(\mathbf S_{p,i}\) is the surface-intersection point. At \(\mathbf S_{p,i}\),
Snell's law gives the water-side ray. With
\(\eta=n_{\mathrm{air}}/n_{\mathrm{water}}\) and interface normal
\(\mathbf n_{\Pi,i}\) oriented against the incident ray,
\begin{equation}
 \begin{aligned}
 c_i&=-\mathbf n_{\Pi,i}^{\mathsf T}\mathbf r_{p,i}^{\mathrm{a}},\quad
 \kappa_i=1-\eta^2(1-c_i^2),\\
 \mathbf r_{p,i}^{\mathrm{w}}&=\eta\mathbf r_{p,i}^{\mathrm{a}}+
 \left(\eta c_i-\sqrt{\kappa_i}\right)\mathbf n_{\Pi,i}.
 \end{aligned}
 \label{eq:snell_ray}
\end{equation}
\(c_i\) is the cosine of the incidence angle, \(\kappa_i\) is the refraction
discriminant, and \(\mathbf r_{p,i}^{\mathrm{w}}\) is the pool-frame,
water-side ray direction.
Thus, each image corner defines a refracted underwater ray
\(\mathbf X_{p,i}(\lambda)=\mathbf S_{p,i}+
\lambda\mathbf r_{p,i}^{\mathrm{w}}\). Notably, our implementation also rejects near-parallel surface intersections,
non-real refracted rays, and rays directed away from the water volume.

\textbf{Pressure Sensor \& IMU.}
Pressure depth and IMU roll/pitch determine the height \(z_i\) of each tag
corner. Intersecting the refracted ray with the horizontal plane \(z=z_i\)
yields the refractively reconstructed 3D corner measurement
\begin{equation}
 \mathbf Q_{p,i}=\mathbf S_{p,i}+
 \frac{z_i-\mathbf e_3^{\mathsf T}\mathbf S_{p,i}}
 {\mathbf e_3^{\mathsf T}\mathbf r_{p,i}^{\mathrm{w}}}
 \mathbf r_{p,i}^{\mathrm{w}},
 \label{eq:refractive_corner_measurement}
\end{equation}
where \(\mathbf e_3=[0,0,1]^{\mathsf T}\) selects the vertical coordinate.

The unknown variables are
\(\boldsymbol{\xi}=[x,y,\psi]^{\mathsf T}\). For a candidate
\(\boldsymbol{\xi}\), the body-to-pool transform \(\mathbf T_{pb}(\boldsymbol{\xi})\)
has rotation and translation, which is given by
\[
\mathbf R_{pb}=\mathbf R_z(\psi)\mathbf R_y(\theta_{\mathrm{imu}})
\mathbf R_x(\phi_{\mathrm{imu}}), \quad \mathbf t_{pb}=[x,y,z_b]^{\mathsf T}.
\]

Rigid-body kinematics predicts the
same corner as
\begin{equation}
 \mathbf P_{p,i}(\boldsymbol{\xi})=
 \mathbf R_{pb}\mathbf P_{b,i}+\mathbf t_{pb}.
 \label{eq:rigid_corner_prediction}
\end{equation}

For the FinsROV implementation, two calibrated AprilTags are mounted on the
vehicle. Let \(\mathcal I\) denote the set of valid corner correspondences from
the detected tags, with \(|\mathcal I|\leq 8\). The refractively reconstructed
corner measurements are aligned with their rigid-body predictions by solving
\begin{equation}
\boldsymbol{\xi}^{*}
=
\arg\min_{x,y,\psi}
\sum_{i\in\mathcal I}
\left\lVert
\mathbf Q_{p,i}
-
\mathbf P_{p,i}(\boldsymbol{\xi})
\right\rVert_2^2.
\label{eq:refractive_pose}
\end{equation}

\begin{figure}[t]
  \centering
  \resizebox{0.95\columnwidth}{!}{\input{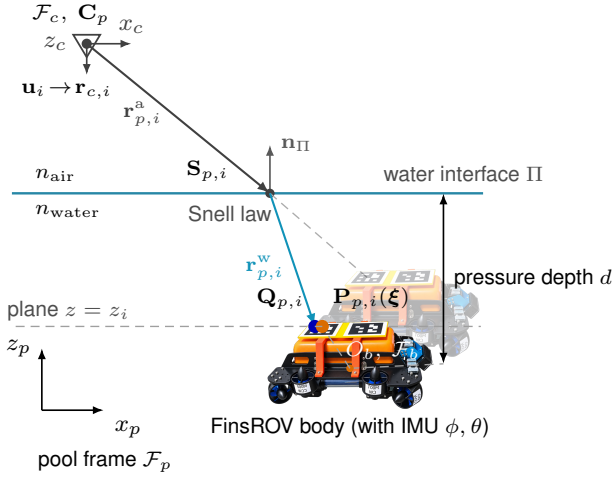}}
  \caption{Constrained refractive AprilTag measurement. A fixed camera observes
  tag corners through a calibrated air-water interface. The production
  underwater measurement optimizes \(x,y,\psi\), with pressure-derived body
  height and IMU roll/pitch as constraints.}
  \label{fig:refractive_geometry}
\end{figure}

A lightweight Levenberg--Marquardt-style
solver~\cite{Marquardt1963NonlinearLeastSquares} computes this constrained
measurement from AprilTag detections~\cite{Olson2011AprilTag}. The multi-tag
redundancy preserves geometric constraints when water-surface disturbances,
glare, or occlusion prevent one tag from being detected.
See Fig.~\ref{fig:refractive_geometry} for details.

\textbf{Multi-Sensor Fusion.}
The resulting measurement is fused by a position-velocity EKF and a separate
yaw filter. Specifically, we define the constant-velocity model over \(\Delta t\) by
\begin{equation}
\label{eq:position_ekf_prediction}
\begin{aligned}
\mathbf A_k&=\begin{bmatrix}
\mathbf I_3 & \Delta t\,\mathbf I_3\\
\mathbf 0_3 & \mathbf I_3
\end{bmatrix},
\quad\hat{\mathbf x}_{k}^{-}=\mathbf A_k\hat{\mathbf x}_{k-1},\\
\boldsymbol{\Sigma}_k^{-}&=\mathbf A_k\boldsymbol{\Sigma}_{k-1}
\mathbf A_k^{\mathsf T}+\mathbf Q_k,
\end{aligned}
\end{equation}
where
\(\mathbf x=[p_x,p_y,p_z,v_x,v_y,v_z]^{\mathsf T}\) denotes the translational state, which contains pool-frame
position \(\mathbf p\) and velocity \(\mathbf v\).

Yaw is propagated from the IMU measurements and corrected by the refractive yaw measurement, which is given by
\begin{equation}
\begin{aligned}
\psi_k^{-}&=\psi_{k-1}+\omega_{z,k}\Delta t,\\
\psi_k&=
\psi_k^{-}+K_{\psi}(\psi_{\mathrm{vis}}-\psi_k^{-}).
\end{aligned}
\label{eq:yaw_ekf}
\end{equation}
where \(\psi_{\mathrm{vis}}\) is the
refractive yaw measurement.
Each visual, depth, and yaw update is accepted only if its squared Mahalanobis
distance satisfies
\(\boldsymbol{\rho}_k^{\mathsf T}
\boldsymbol{\Sigma}_{\rho,k}^{-1}
\boldsymbol{\rho}_k \leq \gamma\),
where \(\boldsymbol{\rho}_k\), \(\boldsymbol{\Sigma}_{\rho,k}\), and \(\gamma\)
denote the innovation, its covariance, and the gating threshold, respectively.

\subsubsection{Actuator Calibration \& Motor-Speed Control}
First, the steady-state thrust curve of each thruster is approximated under
quiescent-water conditions by a direction-dependent quadratic model, which is given by
\begin{equation}
T_i=c_i^{\pm}\omega_i|\omega_i|,
\label{eq:thrust_curve}
\end{equation}
where \(c_i^{+}\) and \(c_i^{-}\) are separately identified for positive and
negative rotation.
Then, the bridge converts a requested force into a target speed and the MCU
closes a low-level speed loop using measured RPM feedback. The implemented
controller is a feedforward-plus-PID form. Notably, the motor model is approximated by the first-order-plus-dead-time (FOPDT) model with an input dead zone.

\subsubsection{Controller to real vehicle}
The controller module supports two control modes. The first, termed \textbf{Thruster8}, directly commands the eight individual thrusters with requested thrust or rotational-speed commands.
The second, termed \textbf{Wrench6}, specifies a desired 6-DOF body-frame
wrench:
\begin{equation}
\boldsymbol{\tau}_d
=
[F_x,F_y,F_z,M_x,M_y,M_z]^{\mathsf T}.
\label{eq:wrench}
\end{equation}
which may be generated by a PID controller, a learned wrench policy, or joystick input. The
allocator then solves the following optimization problem
\begin{equation}
\mathbf f^{*}
=
\arg\min_{\mathbf f^{-}\leq\mathbf f\leq\mathbf f^{+}}
\left\lVert
\mathbf W(\mathbf B\mathbf f-\boldsymbol{\tau}_d)
\right\rVert_2^2
+
\lambda\left\lVert\mathbf f\right\rVert_2^2,
\label{eq:bounded_allocation}
\end{equation}
where \(\mathbf f\in\mathbb{R}^{8}\) is the thrust vector,
\(\mathbf B\in\mathbb{R}^{6\times8}\) is the allocation matrix,
\(\mathbf W\) weights wrench errors, \(\lambda>0\) regularizes thrust effort,
and \(\mathbf f^{-}\) and \(\mathbf f^{+}\) are the thrust bounds.

\section{Experiments}
\label{sec:experiments}

\begin{figure}[t]
  \centering
  \includegraphics[width=0.8\columnwidth]
  {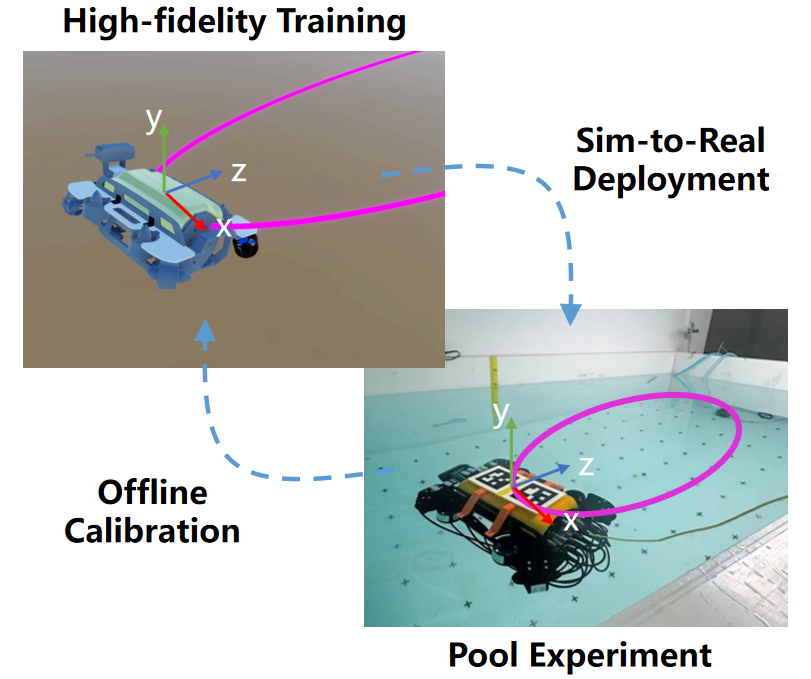}
  \caption{FinsROV pool-experiment setup and the pool world coordinate frame.
  The fixed overhead camera observes the vehicle-mounted AprilTags, while the
  ROV receives depth, inertial, and propulsion telemetry.}
  \label{fig:pool_coordinate}
\end{figure}

As shown in Fig.~\ref{fig:pool_coordinate}, all experiments use FinsROV in the pool setup and pool-fixed
coordinate frame.

\subsection{Localization Reliability}
\label{sec:localization_results}

To isolate the effect of refraction modeling, we evaluate horizontal
localization using \(720\) archived images of a submerged AprilTag grid
captured at \(170\) surveyed locations in the pool. Repeated estimates at each
location are averaged before computing the localization errors. Table~
\ref{tab:localization_accuracy} reports the RMSE and 95th-percentile (P95)
error.

\begin{table}[t]
  \centering
  \caption{Horizontal localization accuracy in the pool.}
  \label{tab:localization_accuracy}
  \footnotesize
  \setlength{\tabcolsep}{7pt}
  \renewcommand{\arraystretch}{1.10}
  \begin{tabular}{@{}lcc@{}}
    \toprule
    Method & RMSE [cm] \(\downarrow\) & P95 [cm] \(\downarrow\) \\
    \midrule
    Pinhole PnP              & 5.74 & 11.41 \\
    Refraction-aware (ours)  & \textbf{0.73} & \textbf{1.45} \\
    \bottomrule
  \end{tabular}
\end{table}

Compared with pinhole PnP, the refraction-aware method reduces both errors by
approximately \(87.3\%\), showing that refraction modeling substantially
improves horizontal localization accuracy in our pool setup.

\subsection{Simulation Dynamics Reliability}
\label{sec:dynamics_results}

\begin{figure}[t]
  \centering
  \includegraphics[width=0.9\columnwidth]{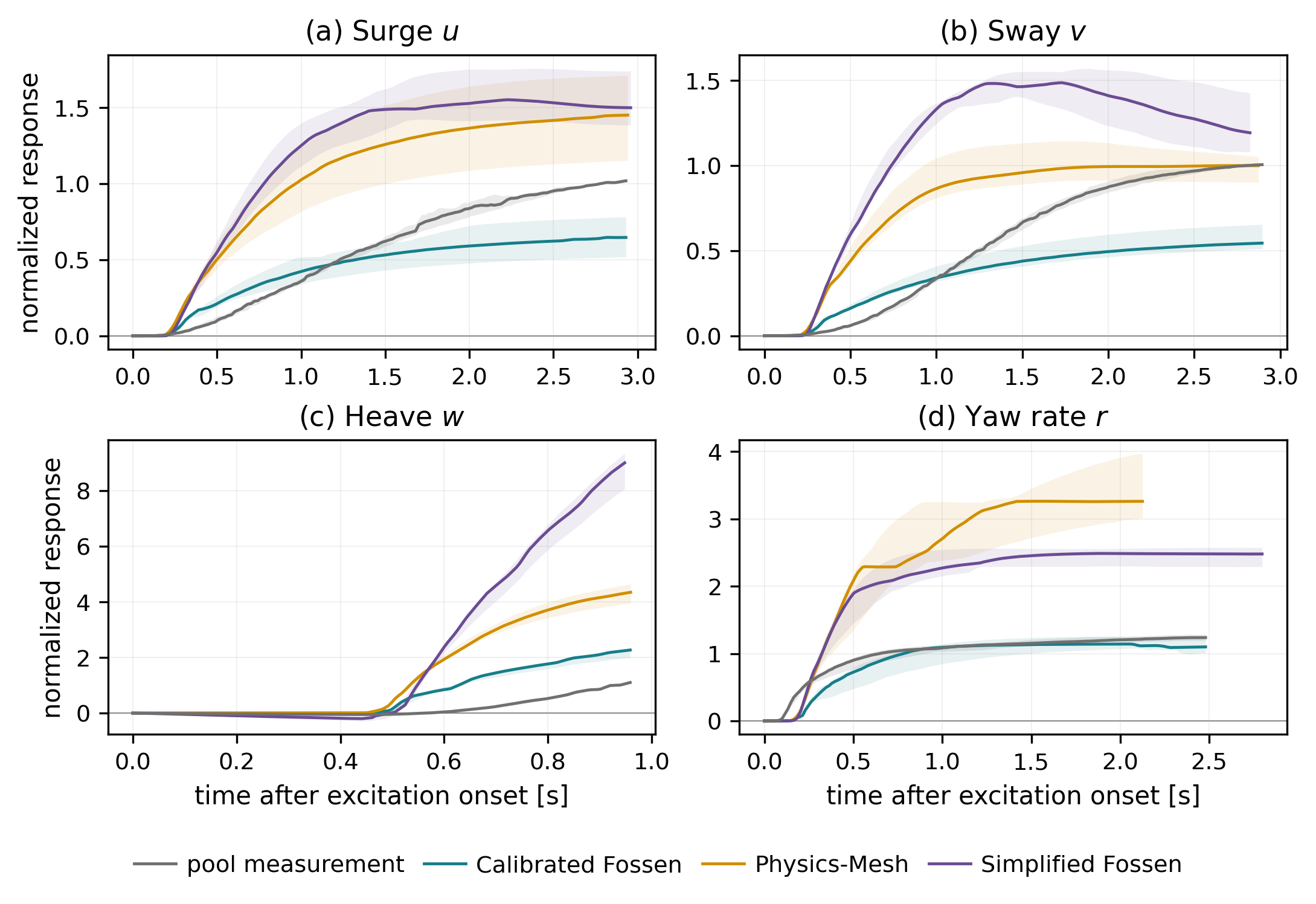}
  \caption{Axis-wise pool-Unity response envelopes for the three simulation
  backends. The responses in each trial are sign-aligned and normalized by the
  5th-95th percentile span of the measured pool response. Solid curves and
  shaded bands denote the pointwise median and interquartile range across
  trials, respectively; the vertical axis represents normalized response magnitude.}
  \label{fig:axiswise_replay_response}
\end{figure}

To evaluate whether FinsSim captures vehicle dynamics rather than merely providing
a task environment, recorded single-axis pool excitations are replayed in Unity
with its three hydrodynamic backends: Calibrated Fossen, Physics-Mesh, and
Simplified Fossen. Simulated responses are compared with the corresponding
physical measurements to assess the dynamic fidelity of each backend.
Fig.~\ref{fig:axiswise_replay_response} shows the response envelopes of \(8\) repeated experiments, while
Table~\ref{tab:dynamics_replay} reports the equal-weight four-axis NRMSEs with
trial-level bootstrap 95\% CIs.
Calibrated Fossen yields the lowest replay error and the closest match to the
physical responses, followed by Physics-Mesh and Simplified Fossen, which is consistent
with the station-keeping transfer results (T1) in
Sec.~\ref{sec:transfer_results}.

\begin{table}[t]
  \centering
  \caption{Four-axis replay errors for the three simulation backends.}
  \label{tab:dynamics_replay}
  \footnotesize
  \setlength{\tabcolsep}{7pt}
  \renewcommand{\arraystretch}{1.08}
  \begin{tabular}{@{}lcc@{}}
    \toprule
    Backend & NRMSE \(\downarrow\) & 95\% CI \\
    \midrule
    Calibrated Fossen & \textbf{0.315} & [0.255, 0.338] \\
    Physics-Mesh      & 1.104          & [0.923, 1.282] \\
    Simplified Fossen & 1.374          & [1.201, 1.447] \\
    \bottomrule
  \end{tabular}
\end{table}

\begin{table*}[t]
  \centering
  \caption{T1 hardware component-ablation and baseline results.}
  \label{tab:t1_hardware_ablation}
  \footnotesize
  \setlength{\tabcolsep}{2.2pt}
  \renewcommand{\arraystretch}{0.98}
  \begin{tabular}{p{6.6cm} c c c c c c c}
    \toprule
    Configuration &
    \(e_{p,\mathrm{3D}}\) [m] & \(e_y\) [m] & \(e_\psi\) [$^\circ$] &
    Success & \(t_s\) [s] & \(J_p\) [m] & \(J_\psi\) [$^\circ$] \\
    \midrule
    \textbf{Calibrated Fossen + NoDR}
      & 0.077 & 0.027 & 6.54 & 7/8 & 20.16 & 0.032 & 2.35 \\
    + DR
      & 0.092 & 0.043 & 5.42 & 7/8 & 36.92 & 0.034 & 2.06 \\
    + localization noise$^\dagger$
      & 0.658 & 0.079 & 13.53 & 2/8 & 41.16 & 0.025 & 3.98 \\
    Without EKF input$^\ddagger$
      & 0.069 & 0.025 & 13.48 & 7/8 & 29.52 & 0.035 & 2.60 \\
    Without first-order thruster dynamics
      & 0.055 & 0.030 & 3.82 & 6/8 & 24.41 & 0.034 & 2.12 \\
    \midrule
    Physics-Mesh
      & 0.746 & 0.108 & 12.19 & 0/8 & - & - & - \\
    Physics-Mesh + DR
      & 0.197 & 0.103 & 8.62 & 0/8 & - & - & - \\
    Simplified Fossen
      & 0.598 & 0.504 & 23.37 & 0/8 & - & - & - \\
    Simplified Fossen + DR
      & 0.609 & 0.564 & 5.48 & 0/8 & - & - & - \\
    \midrule
    Traditional PID
      & 0.176 & 0.101 & 44.08 & 0/8 & - & - & - \\
    \bottomrule
  \end{tabular}

  \vspace{2pt}
  \parbox{0.98\textwidth}{\scriptsize
  \(t_s\) is the median time from
  hold start to the first satisfaction of the T1 success criterion sustained
  for \(10\)~s, over successful trials. \(J_p\) and \(J_\psi\) are position
  and yaw jitter in the first \(10\)-s success-evidence window. ``-'' denotes
  no successful trial.
  \(^{\dagger}\): Independent Gaussian noise with
  \(\sigma_{x,y,z}=5\)~cm and \(\sigma_\psi=10^\circ\) is added only to the controller
  input. \(^{\ddagger}\): The controller bypasses EKF fusion, while the clean EKF
  state is retained for evaluation.}
\end{table*}

\subsection{Sim-to-Real Transfer Reliability}
\label{sec:transfer_results}

To verify whether the policies learned in the simulation environment can be reliably deployed onto the real‑world platform, we test FinsSim on two sim‑to‑real tasks of progressively increasing difficulty, which are adapted from MarineGym~\cite{Chu2025MarineGym}.
Unless stated otherwise, all
trials share the same configuration and deployment pipeline. All policies are trained with PPO~\cite{Schulman2017PPO} in Stable-Baselines3~\cite{Raffin2021SB3} using seed \(42\)
 (it is reasonable because what we focus on is the actual effect after Sim-to-Real transfer, not the policies), while
DR and NoDR denote training with and without domain randomization, respectively. Due to limited space, we only show the results in \textbf{Wrench6} control mode.

\begin{figure}[t]
  \centering
  \includegraphics[width=\columnwidth]{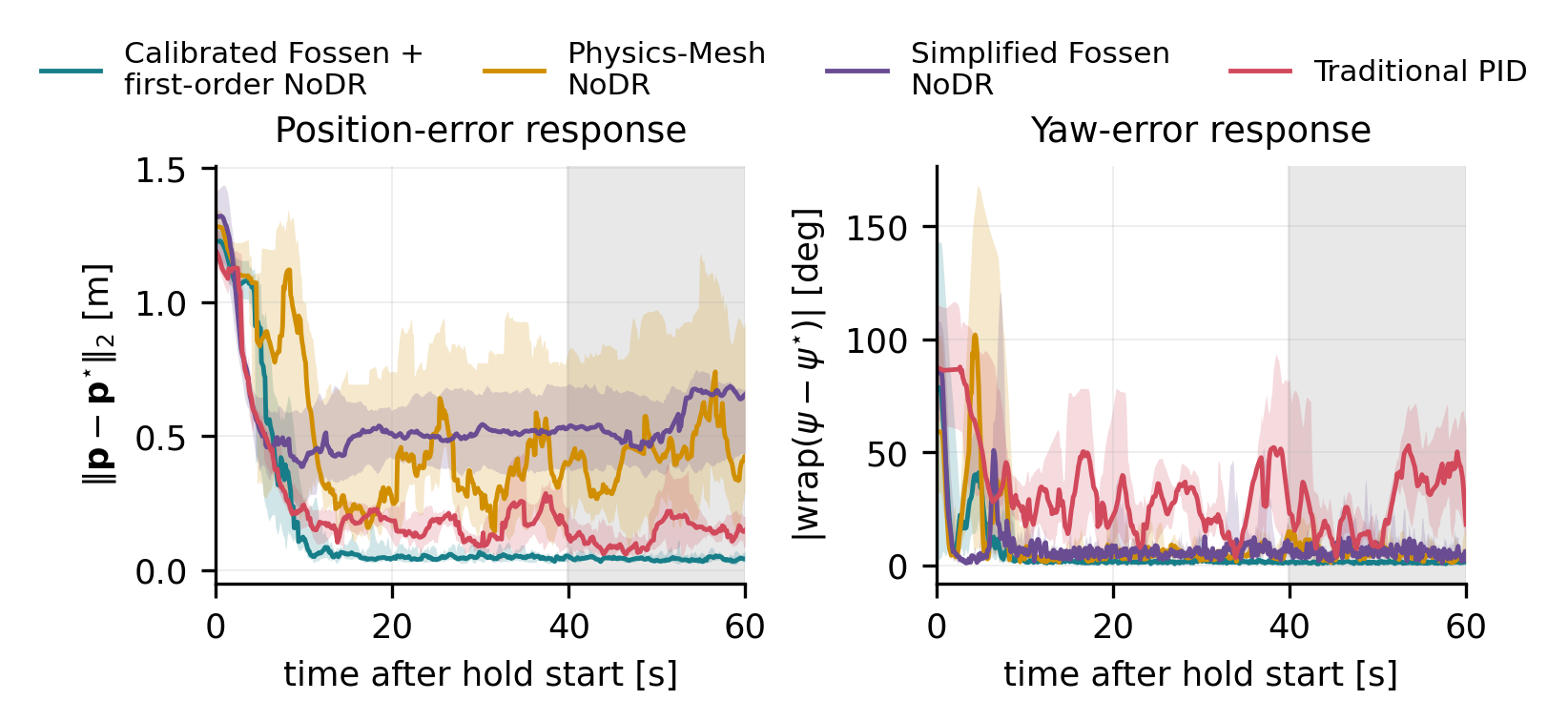}
  \caption{T1 hardware position- and yaw-error envelopes for the three NoDR
  backends and traditional PID. Curves and bands show the trial median and
  interquartile range; the \(40\)-\(60\)~s evaluation window is marked in gray.}
  \label{fig:t1_response_envelopes}
\end{figure}

\textbf{T1: 4-DOF station keeping.}
For a fixed reference
\(\mathbf{p}^{\star}
=[x^{\star},y^{\star},z^{\star},\psi^{\star}]^{\mathsf T}\),
FinsROV regulates eight pre-registered setpoints in \(60\)-s trials, with roll
and pitch stabilized around level. Position and yaw RMSEs are evaluated over
the \(40\)-\(60\)~s late-hold window. Success requires
\(|e_x|,|e_y|,|e_z|\leq0.10\)~m and \(|e_\psi|\leq10^\circ\) continuously for \(10\)~s.
Fig.~\ref{fig:t1_response_envelopes} shows the
error evolution for $8$ setpoints  and Table~\ref{tab:t1_hardware_ablation} reports all tested
configurations.
The Calibrated Fossen reference policy sustains $7$ of $8$ setpoints, with
a 3D position RMSE of \(0.077\)~m and a yaw RMSE of \(6.54^\circ\), whereas
traditional PID sustains none. 

\begin{table}[t]
  \centering
  \caption{T2 hardware tracking errors.}
  \label{tab:t2_results}
  \footnotesize
  \setlength{\tabcolsep}{3pt}
  \renewcommand{\arraystretch}{1.05}
  \begin{tabular}{p{2.5cm} l c c c}
    \toprule
    \textbf{Configuration} & \textbf{Trajectory} &
    \makecell{\textbf{Horizontal}\\\textbf{RMSE [m]}} &
    \makecell{\textbf{Vertical}\\\textbf{RMSE [m]}} &
    \makecell{\textbf{3D P95}\\\textbf{[m]}} \\
    \midrule
    \multirow{4}{2.5cm}{Calibrated Fossen + DR}
      & Straight line & \textbf{0.068} & \textbf{0.129} & 0.409 \\
      & Broken line  & \textbf{0.086} & \textbf{0.084} & \textbf{0.265} \\
      & Ellipse      & \textbf{0.091} & \textbf{0.080} & 0.366 \\
      & Lemniscate   & \textbf{0.051} & \textbf{0.073} & \textbf{0.134} \\
    \midrule
    \multirow{4}{2.5cm}{Calibrated Fossen + NoDR}
      & Straight line & 0.135 & 0.139 & 0.406 \\
      & Broken line  & 0.136 & 0.133 & 0.310 \\
      & Ellipse      & 0.166 & 0.102 & \textbf{0.338} \\
      & Lemniscate   & 0.163 & 0.103 & 0.305 \\
    \midrule
    \multirow{4}{2.5cm}{Traditional PID}
      & Straight line & 0.182 & 0.146 & \textbf{0.405} \\
      & Broken line  & 0.196 & 0.152 & 0.426 \\
      & Ellipse      & 0.174 & 0.134 & 0.433 \\
      & Lemniscate   & 0.207 & 0.170 & 0.514 \\
    \bottomrule
  \end{tabular}
\end{table}
\begin{figure}[htbp]
  \centering
  \includegraphics[width=\columnwidth]{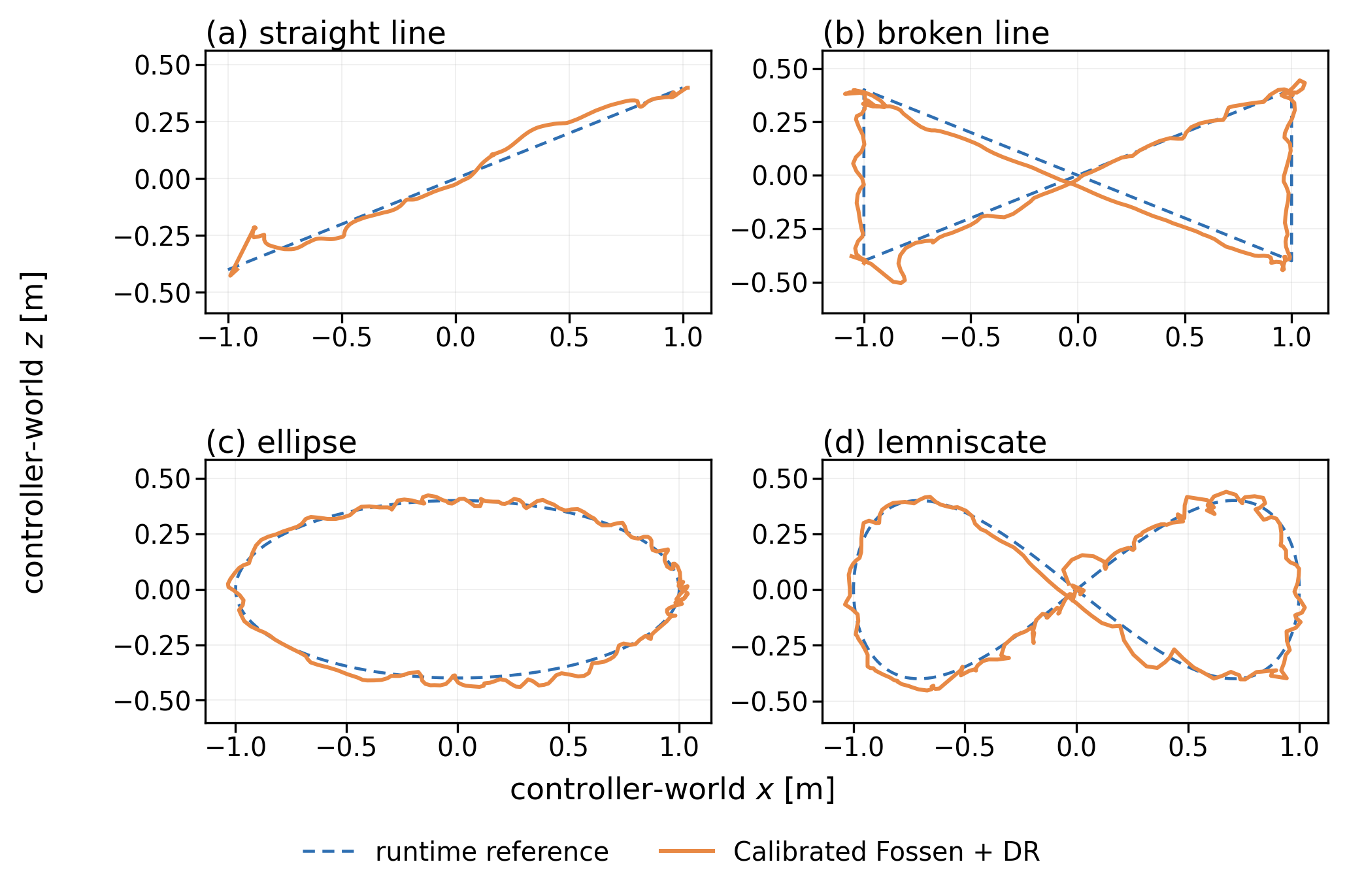}
  \caption{Planar T2 hardware trajectories for Calibrated Fossen + DR. Dashed
  and solid curves denote references and executions. Each panel shows the
  valid trial with the lowest 3D RMSE.}
  \label{fig:t2_fossen_dr_trajectories}
\end{figure}

\textbf{T2: 3D trajectory tracking.}
T2 tracks straight-line, broken-line, elliptical, and lemniscate references
\(\mathbf{p}^{\star}(t)\in\mathbb{R}^{3}\). The \(x\)-\(z\) plane is horizontal
and \(y\) is vertical. Yaw, roll, and pitch are stabilized but not tracked. Fig.~\ref{fig:t2_fossen_dr_trajectories}
shows representative hardware executions. Table~\ref{tab:t2_results} reports horizontal RMSE, vertical RMSE, and the
95th percentile of the 3D position error (3D P95). Among all configurations,
Calibrated Fossen + DR gives lower horizontal and vertical RMSE than PID on all four trajectories and lower 3D P95 on three.

\subsection{Ablation Study}
\label{sec:ablation_study}

To demonstrate the individual effect of each module, we also perform comprehensive ablation experiments.
Specifically, T1 provides a controlled setting for hydrodynamic, sensing, and actuation ablations (see Table~\ref{tab:t1_hardware_ablation}). T2 is used only for the DR comparison (see Table~\ref{tab:t2_results}).

\textbf{Hydrodynamic Backend.}
Without DR, Calibrated Fossen is the only learned configuration that satisfies
the sustained-hold criterion; Physics-Mesh and Simplified Fossen fail at all
eight setpoints. This agrees with Fig.~\ref{fig:t1_response_envelopes} and the order of replay-error in Sec.~\ref{sec:dynamics_results}.

\textbf{Domain Randomization.}
DR has model-, seed- and task-dependent effects, and not always benefits (We conduct experiments on multiple seeds, which are omitted due to limited space).
This is reasonable because DR generates different hydrodynamics and disturbances, which may not be applicable to real environment.
Task complexity also determines the benefit brought by DR.

\textbf{Actuator-Response Model.}
Removing the first-order thruster response reduces the reported errors but
also lowers success count to \(6/8\). Because the calibrated steady-state thrust
curves are unchanged, the transfer benefit of the response model remains inconclusive.

\textbf{Localization and State Estimation.}
Input noise substantially degrades accuracy and success count. Bypassing EKF
fusion preserves similar position RMSE but increases yaw RMSE, indicating
particular sensitivity to yaw-state. This confirms the contribution of the EKF fusion module to robust localization and state estimation.

Overall, the ablations demonstrate that almost
all modules benefit the Sim-to-Real transfer performance.



\section{Conclusions}
\label{sec:conclusions}
In this work, we present
FinsSim, a reality-aligned integrated simulation
platform for underwater robot learning. FinsSim connects Unity and Isaac Lab simulation with reusable learning and control interfaces. The physical platform uses multi-sensor fusion for low-cost localization. Moreover, FinsSim supports Real-to-Sim calibration, which aligns simulation with real-world dynamics and closes the Sim-to-Real loop. Experiments verify the Sim-to-Real reliability of FinsSim. In the future, we expect to continuously optimize FinsSim to make it a more convenient and reliable version, empowering more researches on underwater robot learning.

\section*{Acknowledgment}
The authors used OpenAI ChatGPT and Codex for language polishing, data
organization, figure preparation, and code development. All AI-assisted
materials were carefully reviewed, verified, and approved by the authors, who take full
responsibility for the content and reported results.

\bibliographystyle{IEEEtran}
\bibliography{references}

\end{document}